\documentclass[letterpaper, 10pt, conference]{ieeeconf}
\IEEEoverridecommandlockouts  
\usepackage{amsmath}
\usepackage{algorithm}
\usepackage{algpseudocode}
\usepackage{amsfonts}
\usepackage{graphics} 
\usepackage{graphicx}
\usepackage{amssymb}
\usepackage{caption}
\usepackage{array}
\usepackage{amsmath}
\usepackage{hyperref}
\usepackage{mathtools}
\usepackage{cite}
\usepackage{textcomp}
\usepackage{cleveref}
\usepackage{times}
\usepackage{epsfig}
\usepackage{amsmath}
\usepackage{gensymb}
\usepackage{multirow}
\usepackage{pifont}
\usepackage{subfigure}
\usepackage{url}
\usepackage{tikz}
\usepackage{dblfloatfix}
\usepackage{seqsplit}
\usepackage{booktabs}

\newcommand{\cc}{\color[rgb]{0,0.6,0.3}\checkmark}
\newcommand{\xx}{\color[rgb]{0.6,0,0}{\ding{55}}}
\newcommand{\mmat}[1]{\mathbf{#1}}
\newcommand{\mss}[1]{\mathrm{#1}}

\newcommand{\NSoulCity}{\textit{Soul-city}}
\newcommand{\NSlaughter}{\textit{Slaughter}}
\newcommand{\NJapaneseAlley}{\textit{Japanese-alley}}
\newcommand{\NAutumnForest}{\textit{Autumn-forest}}
\newcommand{\NWinterForest}{\textit{Winter-forest}}
\newcommand{\NOcean}{\textit{Ocean}}
\newcommand{\NFactory}{\textit{Factory}}
\newcommand{\NEndOfWorld}{\textit{End-of-world}}
\newcommand{\NAbandonedFactory}{\textit{Abandoned-factory}}

\begin{document}

\title{TartanAir: A Dataset to Push the Limits of Visual SLAM}

\author{Wenshan Wang$^1$, Delong Zhu$^2$, Xiangwei Wang$^3$, Yaoyu Hu$^1$, \\ Yuheng Qiu$^1$, Chen Wang$^1$, Yafei Hu$^1$, Ashish Kapoor$^4$, Sebastian Scherer$^1$
\thanks{$^1$The authors are with the Robotics Institute of Carnegie Mellon University, Pittsburgh, USA, \textit{email: \{wenshanw, yaoyuh, yuhengq, yafeih, basti\}@andrew.cmu.edu; chenwang@dr.com}
}
\thanks{$^2$The author is with the Department of Electronic Engineering, The Chinese University of Hong Kong, Shatin, N.T., Hong Kong SAR, China. \textit{email: dlzhu@ee.cuhk.edu.hk}
}
\thanks{$^3$The author is with the control science and engineering of Tongji University, Shanghai, China. \textit{email: wangxiangwei.cpp@gmail.com}
}
\thanks{$^4$The author is with Microsoft Research, Seattle, USA. 
\textit{email: akapoor@microsoft.com}
}
}


\maketitle

\begin{abstract}
We present a challenging dataset, the TartanAir, for robot navigation tasks and more. The data is collected in photo-realistic simulation environments with the presence of moving objects, changing light and various weather conditions. By collecting data in simulations, we are able to obtain multi-modal sensor data and precise ground truth labels such as the stereo RGB image, depth image, segmentation, optical flow, camera poses, and LiDAR point cloud. We set up large numbers of environments with various styles and scenes, covering challenging viewpoints and diverse motion patterns that are difficult to achieve by using physical data collection platforms. 
In order to enable data collection at such a large scale, we develop an automatic pipeline, including mapping, trajectory sampling, data processing, and data verification. 
We evaluate the impact of various factors on visual SLAM algorithms using our data. 
The results of state-of-the-art algorithms reveal that the visual SLAM problem is far from solved. Methods that show good performance on established datasets such as KITTI do not perform well in more difficult scenarios. Although we use the simulation, our goal is to push the limits of Visual SLAM algorithms in the real world by providing a challenging benchmark for testing new methods, while also using a large diverse training data for learning-based methods.  
Our dataset is available at \url{http://theairlab.org/tartanair-dataset}. 
\end{abstract}

\section{Introduction}
Simultaneous Localization and Mapping (SLAM) is one of the most fundamental capabilities for robots. 
Due to the ubiquitous availability of images, Visual SLAM (V-SLAM) has become an important component for most autonomous systems~\cite{fuentes2015visual}. Impressive progress has been made with geometric-based \cite{Engel2014lsd, mur2015orb,forster2014svo,engel2017direct},
learning-based~\cite{Zhou2017SfMLearner, wang2018end, wang2019improving}, and hybrid methods  \cite{wang2017non,wang2017noniterative,brahmbhatt2018geometry}.
However, developing robust and reliable SLAM methods for real-world applications remains a challenging problem. Real-life environments are full of inconsistencies such as changing light, low illumination, dynamic objects, and texture-less scenes. 
\begin{figure*} [t]
	\begin{center}
		\includegraphics[width=1.0\textwidth]{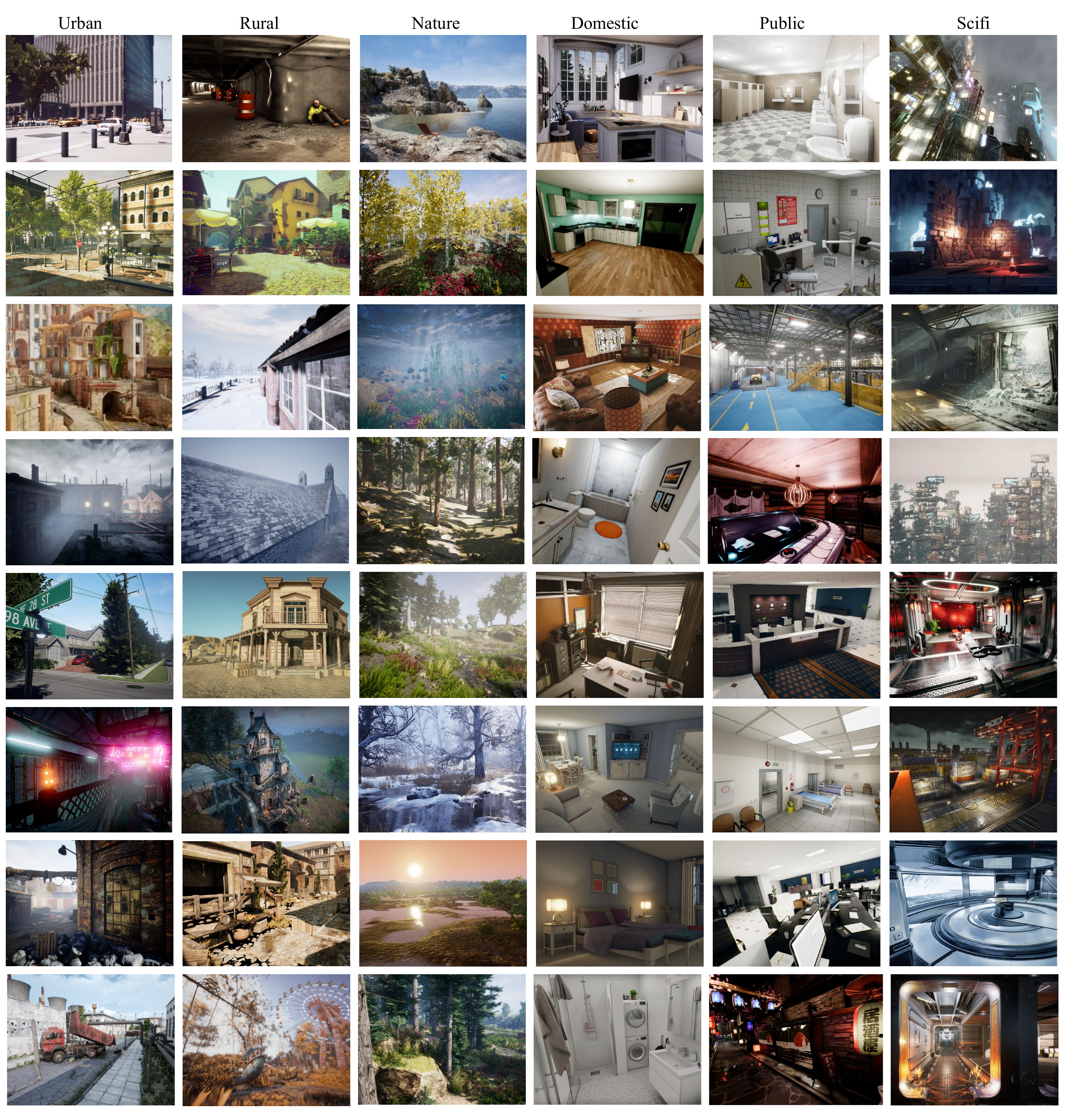}
	\end{center}
	\caption{An overview of the environments. Our dataset is designed to cover a wide range of scenes, which are roughly categorized into urban, rural, nature, domestic, public, and sci-fi. Environments in the same category also have diversity. }
	\label{Fig:envs}
\end{figure*}

The community has been relying heavily on SLAM benchmarks for testing and evaluating their algorithms. On one hand, those benchmarks standardize ways for evaluation, repeatability, and comparison. On the other hand, there is the risk of over-fitting to a benchmark, which means algorithms with a higher score do not necessarily perform better in real-world applications. Current popular benchmarks such as KITTI~\cite{geiger2013vision}, TUM RGB-D SLAM datasets~\cite{sturm12iros} and EuRoC MAV~\cite{Burri25012016} cover generally limited scenarios and motion patterns compared to real-world cases. 

Towards a more challenging benchmark or dataset, \cite{RobotCarDatasetIJRR} presents the RobotCar dataset that contains large scale data in changing light and weather conditions for self-driving tasks. 
Raincouver benchmark focuses on scene parsing tasks in adverse weather and at night~\cite{tung2017raincouver}. KAIST  introduces a multi-spectral dataset covering day/night cases for autonomous and assisted driving \cite{choi2018kaist}. 
However, these datasets, which focus on the self-driving setting, only contain driving scenarios and cover simple motion patterns restricted by the ground vehicle's dynamics. 
On the other hand, The TUM VI dataset~\cite{schubert2018tum} and ScanNet dataset~\cite{dai2017scannet} cover a diverse set of sequences in different scenes but are obtained with a constant light condition in static environments. 


Collecting data in the real world often relies on an elaborate sensor setup and careful calibration. The ground truth for the SLAM/VO task usually comes from other high-accuracy sensors such as LiDAR, GPS, or motion capture system. 
With recent advances in computer graphics, many synthetic datasets have been proposed~\cite{mccormac2017scenenet, gaidon2016virtual,kirsanov2019discoman}. There are trade-offs: the simulation provides reliable ground truth labels, with controllable noise and error; however, one biggest issue known as the sim-to-real gap hampers the algorithm's performance when transferred from the simulation to the real world, due to the distribution difference. On the other hand, physical data collection tools are more difficult to setup. The ground truth is more expensive yet less reliable. Besides, the data distribution is often constrained by the physical property of the hardware, e.g., the data collected by a ground robot often has a fixed roll and pitch angle, most RGB-D cameras are less reliable outdoors, etc.  

To overcome the shortcomings on both sides, we propose to collect a large dataset using photo-realistic simulation environments. Furthermore, we try to minimize sim-to-real gap by increasing diversity. A large number of studies show that by domain randomization \cite{tobin2017domain,tremblay2018training}, namely increasing the diversity of the environment, the model learned in simulation could be easily transferred to the real world. This has been proved to be very effective in many tasks including object detection \cite{tremblay2018training}, robot manipulation \cite{tobin2017domain}, and drone racing \cite{loquercio2019deep}. Our proposed TartanAir dataset is the first such attempt for SLAM-related problems.

In this work, a special focus of our dataset is on challenging environments with changing light condition, low illumination, adverse weather, and dynamic objects. We show in the experiment that state-of-the-art SLAM algorithms struggle in tracking the camera pose in our dataset and frequently get lost on challenging sequences. 
In order to enable large scale data collection, we make a big effort to develop an automatic data collection pipeline, which allows us to process more environments with minimum human intervention.

The contributions of this paper are (1) a large dataset with multi-modal ground truth labels in diverse challenging simulation environments, (2) a fully automatic pipeline for data collection and verification, and (3) we verify the proposed dataset by evaluating popular SLAM algorithms on the dataset, and provide insights on existing problems and future directions of the SLAM algorithms.

\section{Dataset Features}
\label{sec:features}
Although our data is synthetic, we aim to push the SLAM algorithm towards real-world applications. To achieve this, we follow a few design principles. We want a dataset with 1) a large size and high diversity, 2) realistic lighting, 3) multi-modal data and ground truth labels, 4) diversity in motion patterns, and 5) challenging scenarios. 

We adopt the Unreal Engine and collect the data using the AirSim plugin developed by Microsoft \cite{airsim2017fsr}. The Unreal Engine is designed to deliver photo-realistic rendered 3D scenes with complex geometry, high-fidelity texture, dynamic lighting and object motions. Collecting data in simulation allows us to gather a much wider range of appearances, sizes, and motion diversity. In particular, we have found that traditional datasets have limited utility for learning-based methods because they exhibit strong motion biases (e.g., car-like motion at fixed pitch angles). We expect that using simulation allows us to achieve better coverage of scenario and motion patterns, e.g., near-collision actions, aggressive rolling and pitching, which are difficult to gather or annotate in the real world. 

\subsection{Large size diverse realistic data}

We have set up 30 environments, which cover a wide range of scenarios in the real world, from structured urban, indoor scenes to unstructured natural environments~\cite{wang2020visual} (Fig.~\ref{Fig:envs}).
We collected a total of 4TB data, which consists of 1037 long motion sequences. Each of the sequences contains 500-4000 data frames associated with ground truth labels, resulting in more than 1 million frames of data for visual SLAM research. The last column of Table~\ref{Tab:dataset_compare} shows that the proposed TartanAir dataset is at least 1 order of magnitude larger than existing datasets. 


\subsection{Ground truth labels}

\begin{figure}[t]
	\centering  
	\includegraphics[width=0.48\textwidth]{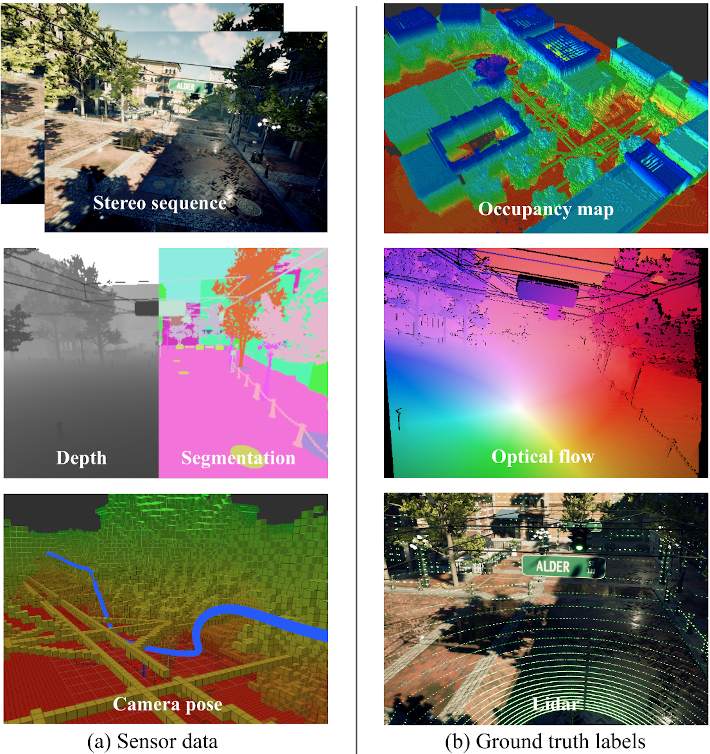}
	\caption{Examples of sensor data and ground truth labels. a) The data provided by the AirSim interface. b) The ground truth labels we calculated using the data. }
	\label{Fig:labels}
	\vspace{-0.2in}
\end{figure}

TartanAir dataset provides multi-modal sensor data and ground truth labels as shown in Table~\ref{Tab:dataset_compare} and Fig.~\ref{Fig:labels}. Using the AirSim interface, we are able to obtain synchronized RGB stereo images, depth images, segmentation labels, and the corresponding camera poses. Based on these data, we developed automatic tools that can generate ground truth occupancy grid maps, optical flow, stereo disparity, and simulated LiDAR measurements. Section~\ref{sec:method} is dedicated to describing the details of the data acquisition. By providing the large-size and multi-modal data, we enable research of visual SLAM in multiple settings, including monocular SLAM, stereo SLAM, RGB-D SLAM, visual LiDAR SLAM, etc. The data can also facilitate a wide range of other visual tasks such as stereo matching, optical flow, monocular depth estimation, and benefit the research in the multi-modal community.

\begin{table*}
	\small
	\begin{center}
		\begin{tabular}{l|ccccc|cccc|c|c}
			\hline
			\multicolumn{1}{c|}{\multirow{2}{*}{Dataset}}                    & \multicolumn{5}{c|}{Sensors/Ground Truth Label}                                                                                                                                                     & \multicolumn{4}{c|}{Conditions}                                                                                         & \multicolumn{1}{l|}{\multirow{2}{*}{\begin{tabular}[c]{@{}l@{}}Motion \\ Pattern\end{tabular}}} & \multirow{2}{*}{\begin{tabular}[c]{@{}l@{}}Seq \\ Num\end{tabular}} \\ \cline{2-11}
			\multicolumn{1}{c|}{}                                            & \multicolumn{1}{l|}{Stereo} & \multicolumn{1}{l|}{Flow} & \multicolumn{1}{l|}{Depth} & \multicolumn{1}{l|}{Lidar} & \multicolumn{1}{l|}{Pose} & \multicolumn{1}{l|}{Light} & \multicolumn{1}{l|}{Weather} & \multicolumn{1}{l|}{Season} & \multicolumn{1}{l|}{DynObj} & \multicolumn{1}{l|}{}                                                                           &                                                                      \\ \hline
			KITTI \cite{geiger2013vision}   		& \cc & \cc & \cc & \cc & \cc & \xx & \xx & \xx & \cc & Car  & 22  \\
			Virtual KITTI \cite{gaidon2016virtual}  & \cc & \cc & \cc & \xx & \cc & \xx & \cc & \xx & \cc & Car  & 50  \\
			EuRoC MAV \cite{Burri25012016}          & \cc & \xx & \xx & \xx & \cc & \cc & \xx & \xx & \xx & MAV  & 11  \\
			TUM RGB-D \cite{sturm12iros}            & \xx & \xx & \cc & \xx & \cc & \xx & \xx & \xx & \cc & Hand  & 15  \\
			ICL-NUM \cite{handa:etal:ICRA2014}      & \xx & \xx & \cc & \xx & \cc & \xx & \xx & \xx & \xx &  Hand & 8   \\
			SceneNet\cite{mccormac2017scenenet}     & \xx & \cc & \cc & \xx & \cc & \cc & \xx & \xx & \xx & Random  & 16K \\
			RobotCar \cite{RobotCarDatasetIJRR}     & \cc & \cc & \cc & \xx & \xx & \cc & \cc & \cc & \cc & Car & - \\
			North Campus \cite{ncarlevaris-2015a}   & \cc & \cc & \cc & \xx & \xx & \xx & \cc & \xx & \cc & Robot  & - \\
			DISCOMAN \cite{kirsanov2019discoman}    & \cc & \cc & \cc & \xx & \xx & \xx & \cc & \xx & \cc & Robot  & 200 \\
			OURS                                    & \cc & \cc & \cc & \cc & \cc & \cc & \cc & \cc & \cc & Random & 1037 \\ \hline
		\end{tabular}
	\vspace{-0.1in}
	\end{center}
	\caption{Comparison of SLAM datasets on sensor data, ground truth labels, scene diversity, motion diversity, and size.}
	\vspace{-0.1in}
	\label{Tab:dataset_compare}
\end{table*}

\begin{figure}[t]
	\centering  
	\subfigure[KITTI Dataset]{\label{fig:model_test_kitti_06}
		\includegraphics[width=0.48\textwidth]{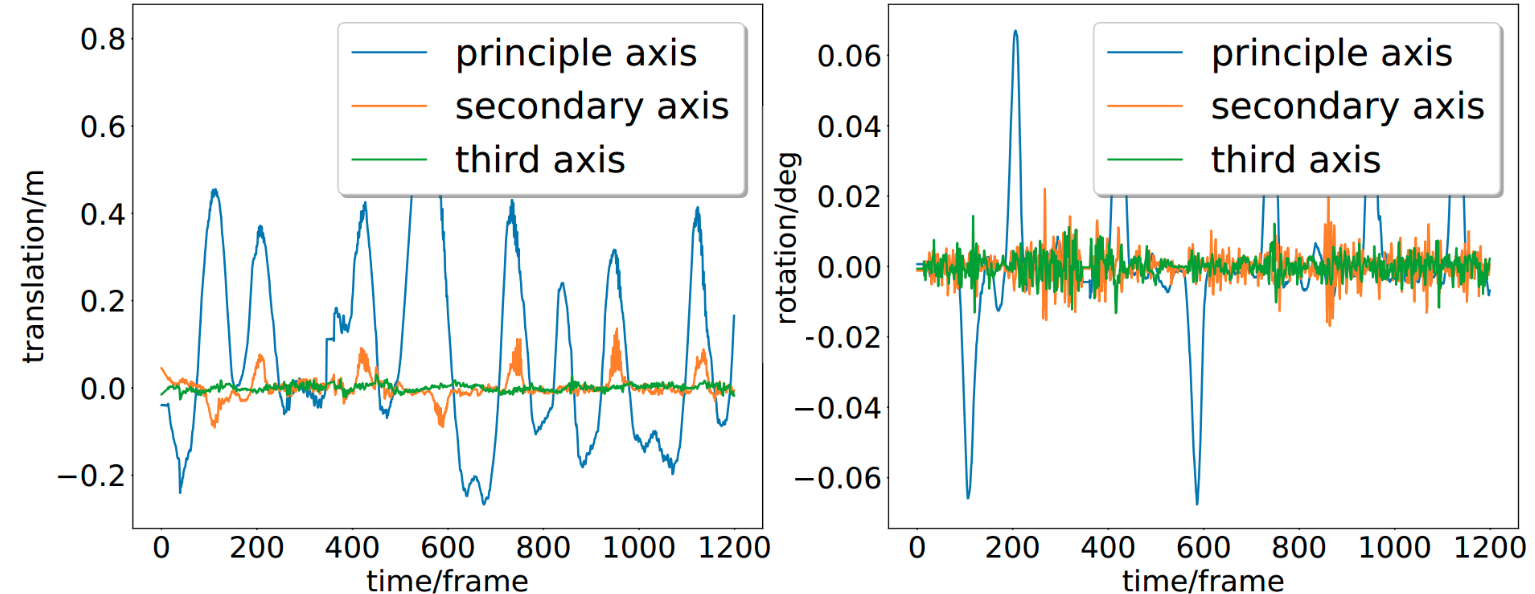}
	}
	\subfigure[Our Dataset]{\label{fig:model_ours_06}
		\includegraphics[width=0.48\textwidth]{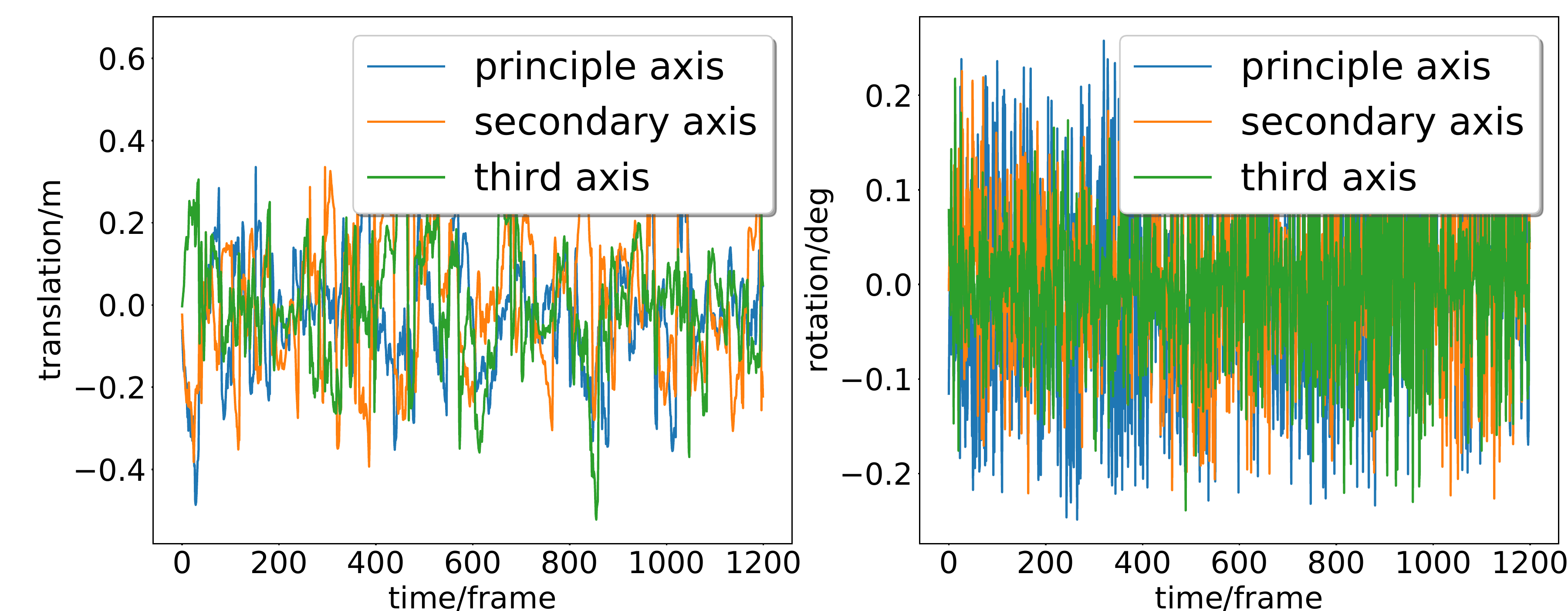}
	} 
	\vspace{-0.1in}
	
	\caption{Visualization of the motion pattern of one sequence from KITTI and TartanAir. The result shows that KITTI has one dominant axis in rotation and translation motion, while the motion pattern in our dataset is more diverse. }
	\label{Fig:motion_pattern}
\end{figure}

\subsection{Diversity of motion patterns}

The existing popular datasets for SLAM such as KITTI~\cite{geiger2013vision} and RobotCar~\cite{RobotCarDatasetIJRR} have a very limited motion pattern, which is mostly moving straight forward combined with small left or right turns.
This regular motion pattern has two limitations. First, the simple motion is insufficient for evaluating a visual SLAM algorithm. We demonstrate in the experiment section that as we increase the complexity of motion patterns, the performance of SLAM algorithms drops significantly.  
Second, learning-based algorithms trained on these data cannot generalize to other tasks with different motion patterns and thus becoming biased.

We randomize the motion distributions and combinations, in order to cover diverse motion patterns in 3D space. We compare the motion patterns between the KITTI dataset and our dataset in Fig.~\ref{Fig:motion_pattern} using Principle Components Analysis (PCA). 
We compute the principle motion components from the translation sequence $\mathbf{T}$ and rotation sequence $\mathbf{R}$ respectively, where $\mathbf{T} \in \mathbb{R}^{3\times n}$ concatenates $n$ frames of translation motion $(\Delta x, \Delta y, \Delta z)$,
and $\mathbf{R} \in \mathbb{R}^{3\times n} $ includes $n$ rotation motions in $so(3)$ format.
The principal components of a motion sequence $(\mathbf{T},\mathbf{R})$ could be decomposed in Eq.~\eqref{eq:T} and \eqref{eq:R}.
\begin{equation} \label{eq:T}
U_{\mss{trans}}\mathrm{diag}(t_1,t_2,t_3) V_{\mss{trans}}^{*} = \mathbf{T}
\end{equation}
\begin{equation} \label{eq:R}
U_{\mss{rot}}\mathrm{diag}(r_1,r_2,r_3) V_{\mss{rot}}^{*} = \mathbf{R}
\end{equation}
where $t_1,t_2,t_3$ and $r_1, r_2, r_3$ represent the principle motion of a given sequence. At the same time, we obtain 3 eigenvectors, which define a new vector space. Then we project each frame to this new vector space and plot the values in Fig.~\ref{Fig:motion_pattern}. As expected, the data from KITTI has one dominant axis in both translation and rotation. In our case, we are able to achieve more diverse motion patterns.  One can also deliberately add constraints in the simulation (e.g., fix roll and pitch angles), so as to mimic a ground robot’s motion pattern.

Furthermore, we propose a new metric to evaluate the diversity of motion patterns. 
We define the motion diversity metric as:
\begin{equation}
\sigma = \frac{1}{2}(\frac{\sqrt{t_2 t_3}}{t_1} + \frac{\sqrt{r_2 r_3}}{r_1}) 
\end{equation}
\begin{table}
	\small
	\begin{center}
		\begin{tabular}{cccccc}
			\hline
			Name & KITTI & EuRoC & TUM & RobotCar & Ours\\
			\hline
			$\sigma$ & 0.005 & 0.207 &  0.196 & 0.047 & 0.95 \\
			\hline
		\end{tabular}
	\end{center}
	\vspace{-0.1in}
	\caption{Comparison of the diversity of the motion pattern under the propose metric. Larger value means the motion is more diverse. }
	\label{Tab:motion_pattern}
	\vspace{-0.2in}
\end{table} 
We present the $\sigma$ of SLAM datasets under this metric in Table \ref{Tab:motion_pattern}. A small $\sigma$ indicates that the motion is dominated by one dimension. $\sigma$ converges to 1 as the diversity of the motion pattern increases. According to our evaluation, the TUM dataset collected with a handheld camera and the EuRoC MAV dataset collected with MAV also get low scores, which indicate the motion pattern is constrained by the MAV dynamics and human habits.  

\begin{figure}[t]
	\centering  
	\includegraphics[width=0.5\textwidth]{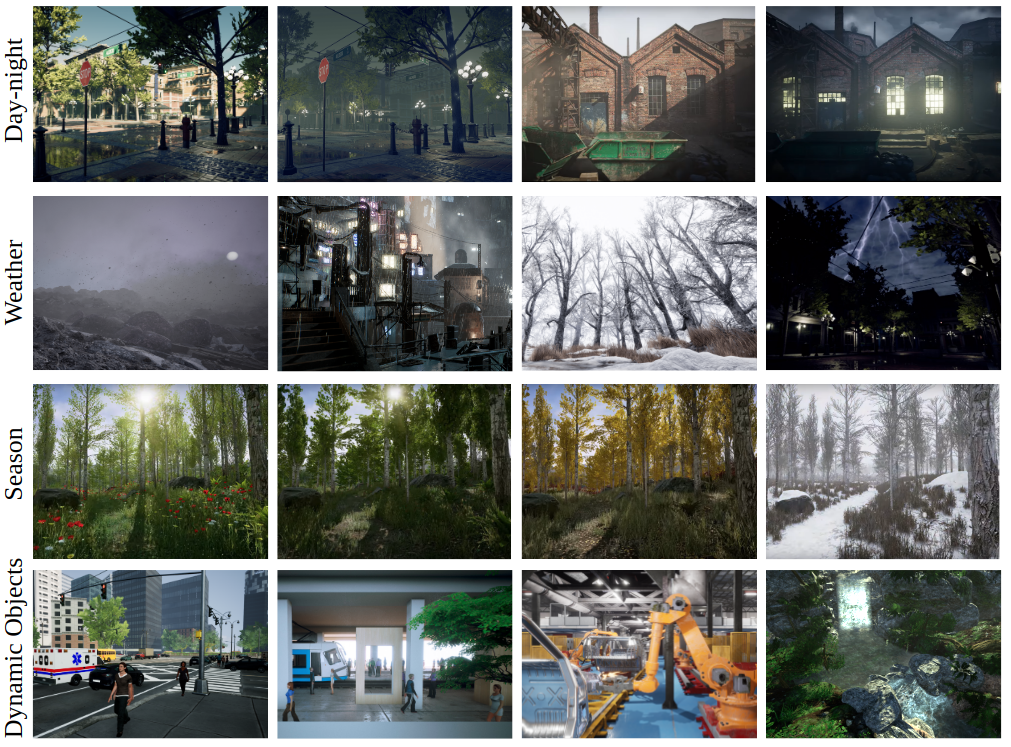}
	\caption{Challenging scenarios.}
	\label{Fig:challenging}
	\vspace{-0.2in}
\end{figure}

\subsection{Challenging Scenes}

TartanAir contains a large number of challenging environments, including dynamic light conditions, low illumination, adverse weather, and dynamic objects (Fig.~\ref{Fig:challenging}).   

Utilizing the Unreal Engine allows us to render realistic 3D scenes with dynamic lighting. TartanAir covers scenarios with strong lighting changes, shadows, over-exposure, reflections of glass or puddle. The light sources range from road lamps, neon lights to sunlight and moonlight. When collecting data through AirSim, the cameras can be configured to an auto-exposure mode, similar to real cameras. This feature adds another layer of dynamics in response to dynamic lighting.

We augment the outdoor environments with various effects of weather conditions and seasonal changes. We have collected data at different times-of-day and different seasons, while it is raining, snowing, and foggy.

TartanAir contains dynamic objects that consist of simulated humans, vehicles, machinery, dynamic vegetation such as waving leaves and grass. The motion of the objects can be configured to different levels of dynamics. 

In the experiment, we compare the SLAM algorithm with and without these challenging settings. Results show that the SLAM algorithms are heavily affected by these challenging conditions. 

\begin{figure*}[t]
	\centering
	\includegraphics[width=1.0\textwidth]{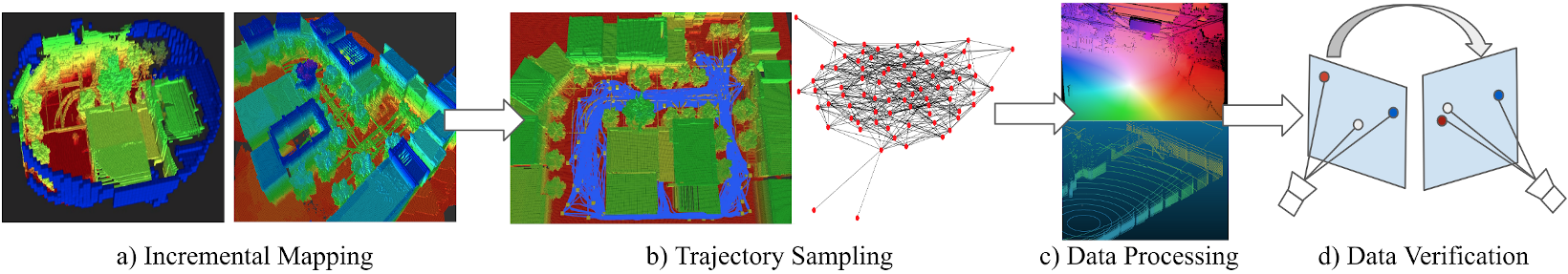}
	\caption{An overview of the data collection pipeline.}
	\label{fig:method_overview}
	\vspace{-0.2in}
\end{figure*}

\section{Method}
\label{sec:method}
To collect the raw data on such a large scale, we design a fully automatic pipeline, as shown in Fig.~\ref{fig:method_overview}, which allows us to easily scale up the collection to 30 diverse environments. The whole pipeline includes four modules which are incremental mapping, trajectory sampling, data processing, and data verification. The mapping process traverses the target environment and reconstructs an accurate occupancy grid map for obstacle avoidance and path planning. Based on the maps, the trajectory sampling process generates a large amount of collision-free trajectories, while maximizing the diversity of viewpoints and motion patterns. We then dictate virtual cameras to move along the trajectories and collect multi-modal sensor data, e.g., camera poses, RGB, segmentation, and depth images. Based on the collected data, we also calculate other ground truth labels including optical flow, LiDAR, and stereo disparity. At last, the data verification module verifies the correctness of the data. The entire system can autonomously run with minimal human interventions, which is the key to enabling a large scale data collection process. 

\subsection{Incremental Mapping}
Incremental mapping refers to the process of actively mapping unknown environments and collecting as much information as possible~\cite{zhu2018rlsuper, srmchaoqun}. The occupancy grid map is a frequently-used map representation method. We utilize the depth image and camera pose as the input of the mapping process, and implement a frontier-based algorithm to automatically calculate the next mapping location (Fig.~\ref{fig:method_overview} a). The RRT* planning algorithm~\cite{noreen2016comparison} is leveraged to plan collision-free trajectories, which navigate the robot to the target mapping locations. The mapping process ends when there are no more frontiers in the region. The time for mapping an environment with a size of 100m x 100m x 10m at a resolution of 0.25m is about 1 hour.  

\subsection{Trajectory Sampling}
Trajectory sampling consists of a graph generation process and a data collection process.
During the graph generation, we randomly sample N nodes in the free space.  
We then apply RRT* to plan a safe trajectory between each pair of nodes. The nodes and edges are stored in a graph data structure. After a large number of samplings, a trajectory graph that encodes the feasible paths of the environment is generated (Fig.~\ref{fig:method_overview} b). Then we sample loop trajectories from the graph. The trajectories are further processed using spline smoothing techniques while avoiding the obstacles. 
In the data collection process, we randomize the incremental distance and angles along the trajectory. The poses are sent to a virtual camera in the simulation environment, and all the required data is recorded through the AirSim interface.






\subsection{Data Processing}

As discussed previously, camera poses, RGB and depth images, and semantic segmentation labels are directly obtained from the environment. The ground truth data such as optical flow, stereo disparity, and simulated LiDAR measurements are generated from these raw data.

\begin{figure}[ht]
	\centering
	\includegraphics[width=0.48\textwidth]{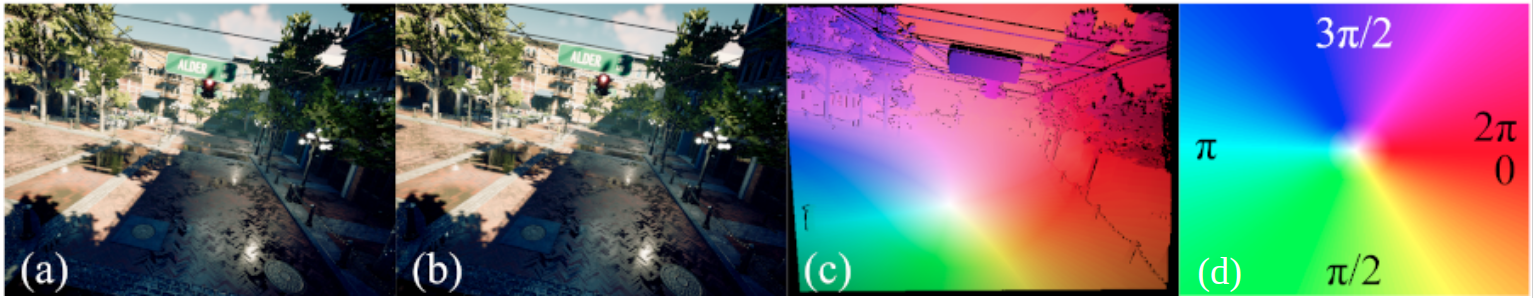}
	\caption{Sample optical flow. (a)(b) Reference and test images. (c) Optical flow. (d) Color mapping.}
	\label{Fig:OptFlowNoMask}
	\vspace{-0.1in}
\end{figure}

\noindent\textbf{Optical flow} The ground truth optical flow is calculated for static environments by image warping. For each pair of camera poses along a trajectory, 
we refer the first camera as reference camera $C^{\mss{ref}}$ and the second as test camera $C^{\mss{tst}}$, 
and define $\mmat{D}^{\mss{ref}}$ and $\mmat{D}^{\mss{tst}}$ as two depth images. Optical flow values are calculated for each pixel by transforming the entire $\mmat{D}^{\mss{ref}}$ from $C^{\mss{ref}}$ to $C^{\mss{tst}}$ according to the camera poses. In practice, we convert $\mmat{D}^{\mss{ref}}$ to a 3D point cloud and project the point cloud to the image plane of $C^{\mss{tst}}$. Let $\left ( x_p, y_p \right )^{\mss{ref}}_{\mss{r}}$ be the image coordinate of a pixel $p$ in $\mmat{C}^{\mss{ref}}$ and $\left ( x_p, y_p \right )^{\mss{tst}}_{\mss{r}}$ be its transformed image coordinate in $\mmat{C}^{\mss{tst}}$. We use subscript $\mss{r}$ to denote an entity originally observed in $\mmat{C}^{\mss{ref}}$. Optical flow is directly measured as $f_p = (f_x, f_y)_p = \left ( x_p, y_p \right )^{\mss{tst}}_{\mss{r}} - \left ( x_p, y_p \right )^{\mss{ref}}_{\mss{r}}$, where $f_p = (f_x, f_y)_p$ is the optical flow of $p$ observed in $\mmat{C}^{\mss{ref}}$. 
Fig.~\ref{Fig:OptFlowNoMask} shows a sample optical flow visualized similar to the KITTI~\cite{geiger2013vision} dataset.

\begin{figure}[ht]
	\centering
	\includegraphics[width=0.48\textwidth]{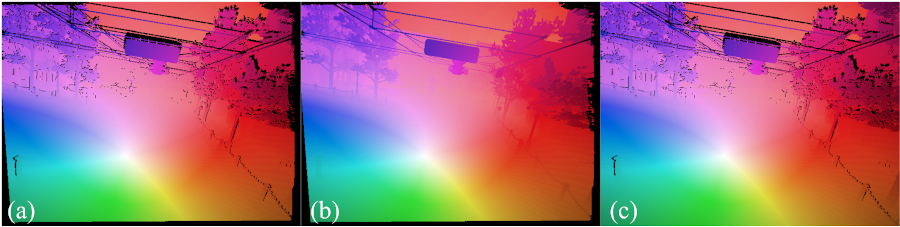}
	\caption{Optical flow with masks. (a) Both masks. (b) Out-of-FOV mask only. (c) Occlusion mask only.}
	\label{Fig:OptFlowMask}
	\vspace{-0.2in}
\end{figure}

We also provide two masks over the optical flow image (Fig.~\ref{Fig:OptFlowMask}): the occlusion mask and the out-of-FOV (field of view) mask. 
We check occlusion for each pixel considering the change of camera pose and obstacles that are only visible in $C^{\mss{tst}}$. The out-of-FOV mask records all the pixels in $\mmat{D}^{\mss{ref}}_{\mss{r}}$ which fall out of the FOV of $C^{\mss{tst}}$. 
Objects observed in $C^{\mss{ref}}$ may be located behind the camera center of $C^{\mss{tst}}$. Pixels in such cases are labeled as invalid and are also saved in the out-of-FOV mask.

\begin{figure}[ht]
	\centering
	\includegraphics[width=0.48\textwidth]{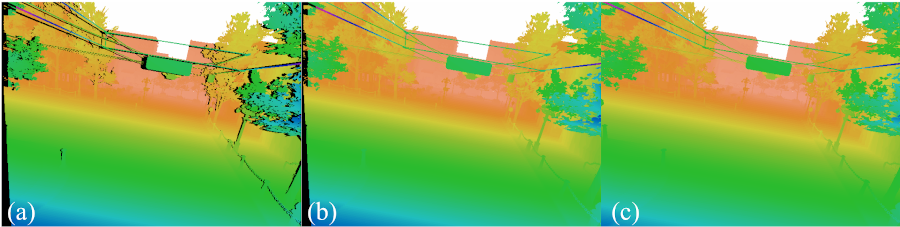}
	\caption{Stereo disparity with masks. (a) Both masks. (b) Out-of-FOV mask only. (c) No masks. Color: Purplish and dark pixels have large disparity, reddish and bright pixels have small disparity, masked pixels are black.}
	\label{Fig:StereoDisparityMask}
	\vspace{-0.1in}
\end{figure}

\noindent\textbf{Disparity} The disparity value is calculated from the depth image and the camera intrinsic value. We also calculate the disparity mask similar to the optical flow calculation. Similar to optical flow, we produce the occlusion mask and the out-of-FOV mask. The disparity images correspond to Fig.~\ref{Fig:OptFlowNoMask} (a) are shown in Fig.~\ref{Fig:StereoDisparityMask}.  

\begin{figure}[ht]
	\centering
	\includegraphics[width=0.48\textwidth]{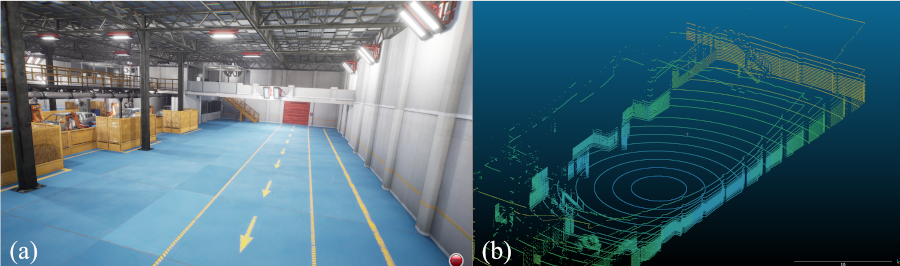}
	\caption{Extracted LiDAR points. (a) Virtual environment. (b) 32-line LiDAR points. }
	\label{Fig:LiDAR}
	\vspace{-0.2in}
\end{figure}

\noindent\textbf{LiDAR} We extract LiDAR points by sampling depth value from a virtual camera and mimicking a LiDAR device. We put 4 90$\degree$-FOV cameras at the same spatial position with a resolution such that every extracted LiDAR point does not share a same pixel with its neighbor LiDAR points. The distance of a LiDAR point is interpolated over the nearest depth pixels. Fig.~\ref{Fig:LiDAR} shows a LiDAR point cloud by simulating a 32-line LiDAR. 

\subsection{Data Verification}

An issue we encountered in the early stage is that the camera pose and the image are not synchronous. 
We add the pause feature to the AirSim to ensure the synchronization. 
To verify the synchronization, we calculate the optical flow for the consecutive image pair. Then, the pixels of $C^{\mss{ref}}$ are projected to $C^{\mss{tst}}$ according to the optical flow, and the mean photometric error (RGB channels) is evaluated. Due to the viewing angle, lighting, and surface reflection, some projected pixels have large photometric errors. However, the averaged error is roughly constant and small. By monitoring the error we effectively verify the synchronization problem. As an example, the sequence in Fig.~\ref{Fig:OptFlowNoMask} has approximately 700 images. The maximum mean photometric error over all the consecutive image pairs is less than 5 with the maximum possible value being $255\sqrt{3}$. Additionally, we detect images with a large area of occlusion by the optical flow masks. The depth images are used to verify the collision with the environment.

\section{Experimental Evaluation}

\begin{figure}[ht]
	\centering
	\includegraphics[width=0.48\textwidth]{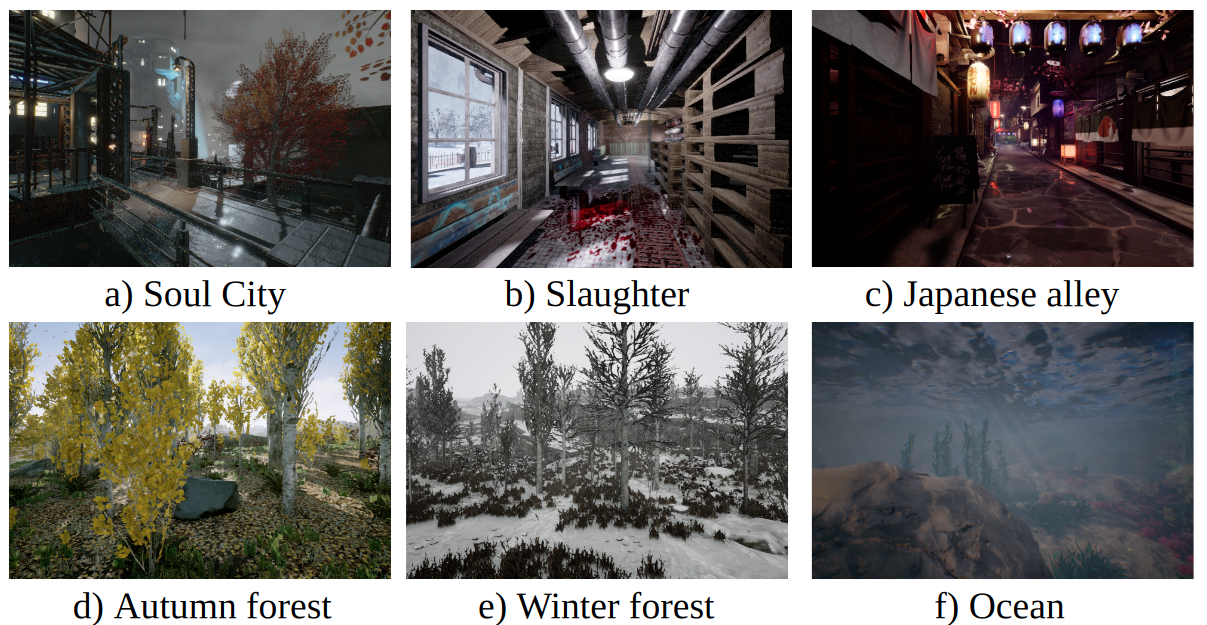}
	\caption{Selected environments for baseline comparison. a) Raining, lens flare. b) Changing light condition (blinking lights). c) Low illumination, indoor/outdoor transition. d) Falling leaves. e) Less feature on the ground. f) Dynamic objects: fish and bubbles. }
	\label{Fig:exp_envs}
	\vspace{-0.2in}
\end{figure}

We show experimental results of applying ORBSLAM-Monocular (ORB-M), ORBSLAM-Stereo (ORB-S) \cite{mur2015orb} and DSO-Monocular \cite{engel2017direct} to 6 representative environments with interesting features, as shown in Fig.~\ref{Fig:exp_envs}. 

\subsection{Testing environments}
The \NSoulCity{} is an outdoor raining environment with camera lens flare effects. The \NSlaughter{} is an indoor scene with challenging blinking lights in some parts of the hallway. The \NJapaneseAlley{} is a night scene switching between indoors and outdoors. The \NAutumnForest{} is a dynamic environment with a lot of falling leaves. The \NWinterForest{} is covered by snow which results in fewer features on the ground. The \NOcean{} environment has fish flock and bubbles moving in the water.

\begin{table*}
	\centering
	\begin{tabular}{l|l|cccc|cccc|cccc}
		\hline
		\multirow{2}{*}{\begin{tabular}[c]{@{}l@{}}Env Name\end{tabular}} & \multirow{2}{*}{\begin{tabular}[c]{@{}l@{}}Motion \\ Pattern\end{tabular}} & \multicolumn{4}{c|}{ORB-M}                   & \multicolumn{4}{c|}{ORB-S}                   & \multicolumn{4}{c}{DSO}   \\ \cline{3-14} 
		&                     & \multicolumn{1}{c}{ATE} & \multicolumn{1}{c}{RPE-T}  & \multicolumn{1}{c}{RPE-R}  & SR      & \multicolumn{1}{c}{ATE} & \multicolumn{1}{c}{RPE-T}  & \multicolumn{1}{c}{RPE-R}  & SR      & \multicolumn{1}{c}{ATE} & \multicolumn{1}{c}{RPE-T}  & \multicolumn{1}{c}{RPE-R}  & SR      \\ \hline
		\multirow{3}{*}{Soul-city}        
           & Easy             & 0.23    & 0.005  &0.361 & 0.48    & 0.131  & 0.015  & 1.130 & \textbf{0.91}    & 0.405   & 0.034  &0.480 & 0.45      \\
           & Mid              & 0.06    & 0.015 & 0.122 & 0.68    & 0.249  & 0.045 & 1.071 & \textbf{0.79}    & 0.957   & 0.053 & 1.215 & 0.39     \\
           & Hard             & 0.03   & 0.002  & 0.073& \textbf{0.25}    & 3.961  & 0.308 & 7.953 & 0.13    & 12.065   & 1.283 &  11.371 & 0.06     \\ \hline
		\multirow{3}{*}{Slaughter}       
           & Easy             & 0.517   & 0.052  & 0.562 & 0.48    & 0.232   & 0.073 & 0.816 & \textbf{0.75}    & 1.699   & 0.440 & 1.976 & 0.22      \\
           & Mid              & 0.112   & 0.018  & 0.183 & 0.26    & 0.430  & 0.191 & 1.431 & \textbf{0.43}   & 1.87  & 0.443  & 3.077 & 0.18      \\
           & Hard             & 0.298  & 0.053  & 0.542 & \textbf{0.10}    & - & - & - & 0   & 13.119   & 1.247  & 8.147 & 0.1      \\ \hline
		\multirow{3}{*}{Japanese-alley}   
           & Easy             & 0.257   & 0.016  & 0.270 & 0.98    & 0.057   & 0.018 & 0.704 & \textbf{1.0}     & 0.195   & 0.027 & 0.300 & 0.69      \\
           & Mid              & 0.449  & 0.048 & 0.691 & 0.636    & 0.223   & 0.071  & 0.791 & 0.77    & 1.247   & 0.117 & 1.279 & \textbf{0.77}  \\
           & Hard             & 0.709   & 0.076  & 0.642& 0.25    & 1.299  & 0.406 & 4.499 & 0.22    & 4.090   & 0.346  & 3.10 & \textbf{0.44}     \\ \hline
		\multirow{3}{*}{Autumn-forest}   
           & Easy             & 0.083   & 0.006 & 0.138 & 0.47    & 0.036   & 0.010  & 0.502 & \textbf{0.99}    & 1.660   & 0.016  & 1.092 & 0.025    \\
           & Mid              & 0.051   & 0.007 &0.076   & 0.28    & 0.192   & 0.032  & 0.672 & \textbf{0.68}    & -      & -   &  -  & 0        \\
           & Hard             & 0.007   & 0.002& 0.014 & \textbf{0.07}    & 1.468  & 0.462 & 3.857   & 0.05    & -      & -   &  -  & 0        \\ \hline
		\multirow{3}{*}{Winter-forest}              
           & Easy             & 0.051   & 0.006 & 0.082 & 0.87   & 0.018   & 0.005 & 0.361 & \textbf{1.0}    & -      & -    & -  & 0        \\
           & Mid              & 0.136   & 0.012 & 0.197 & 0.45   & 0.109   & 0.025  & 0.478 & \textbf{0.87}   & 2.305   & 0.196 & 1.745 & 0.12   \\
           & Hard             & 0.043   & 0.009 & 0.084 & \textbf{0.30}   & 0.772 & 0.137 & 2.683  & 0.18   & 3.707   & 0.870  & 4.566 & 0.09    \\ \hline
		\multirow{3}{*}{Ocean}           
           & Easy             & 0.200   & 0.034&0.448   & 0.78   & 0.465  & 0.067 & 1.844 & \textbf{0.98}   & 2.662   & 0.270 & 3.170 & 0.33   \\
           & Mid              & 0.286  & 0.026 & 0.512 & 0.44   & 0.658   & 0.148  & 2.603 & \textbf{0.76}   & 4.262   & 0.607 & 5.510 & 0.42   \\
           & Hard             & 0.138   & 0.010 & 0.341 & 0.09   & 2.061  & 0.713 & 6.370 & 0.09   & 11.662   & 1.088 & 12.537 & \textbf{0.31}    \\ \hline
	\end{tabular}
	\caption{Comparison of SLAM methods in multiple environments. Bold number shows the best SR for each setting. }
	\vspace{-0.2in}
	\label{Tab:baselines}
\end{table*}

\subsection{Metrics}
We use 3 metrics, namely absolute trajectory error (ATE), relative pose error (RPE) and success rate (SR), to evaluate the algorithms. Because monocular methods cannot recover the absolute scale information, we perform a scale correction before calculating the ATE and RPE for monocular algorithms \cite{geiger2013vision}. The SR is defined as the ratio of non-lost sequences to the number of total sequences.

We find that for challenging datasets as ours, the SR metric better captures the performance of the algorithms. \textit{While ATE and RPE are less reliable because they can only be calculated on successful trajectories}. 
For harder sequences, less robust algorithms fail more often to track the camera poses, thus they do not return ATE and RPE scores. As a result, less robust algorithms could get higher ATE and RPE since they give up hard sequences. 

On the other hand, the SR is affected by the length of the trajectory, since a longer sequence is often more difficult to complete. Consequently, we cut the sequences to the same length of 200 frames. 

\subsection{Evaluation results}
We collect 30-50 sequences from each environment for testing. In addition, we define 3 difficulty settings in terms of motion pattern (Table~\ref{Tab:difficulty}). In the easy mode, pitch and roll angles are fixed, which is similar to a ground robot setting. The medium and hard modes have 6 DoF motion. We increase the maximum translation and rotation speed from easy to hard. 

\begin{table} [ht]
	\begin{center}
		\begin{tabular}{lccc}
			\toprule
			& Motion DoF & MaxTrans (m) & MaxAngle ($\degree$) \\
			\midrule
			Easy & Trans+Yaw & $\pm{0.2}$ & $\pm{3}$\\
			Medium & 6 DoF & $\pm{0.3}$ & $\pm{5}$ \\
			Hard & 6 DoF & $\pm{0.5}$ & $\pm{10}$\\
			\bottomrule
		\end{tabular}
	\end{center}
	\vspace{-0.1in}
	\caption{The settings of 3 difficulty levels. Motion DoF indicates the motion complexity. In the easy mode we fix the pitch and roll rotation. MaxTrans represents the randomized range of translations and MaxAngle represents the randomized range of rotation between  consecutive frames. }
	\vspace{-0.1in}
	\label{Tab:difficulty}
\end{table}

We compare ORB-M, ORB-S, and DSO-M on the aforementioned 3 metrics in 3 difficulty levels. Since ORBSLAM and DSO are non-deterministic, we repeat the experiments 5 times and report the mean value. 
As shown in Table~\ref{Tab:baselines}, the SR drops remarkably as the difficulty of the motion pattern increases. Even with the easy motion pattern, the monocular algorithms have a low SR score, which indicates they suffer from the challenges in the scenes. 
We observe a few interesting outcomes from the experiment. First of all, as expected, the ORB-S is more robust than ORB-M and DSO in all 6 of easy mode cases, 5/6 of medium cases. But ORB-S performs worse than ORB-M in all 6 of hard cases (the difference is small though). The accuracy and robustness of DSO are generally worse, one reason could be it does not have a loop closure. However, DSO performs best in \NJapaneseAlley{}, which is challenging due to low illumination. This reflects the advantage of DSO as a direct method compared to the feature-based method in a low feature environment. 
We find that the failure cases are consistent with those reported in the real world \cite{younes2017keyframe}. The visual SLAM algorithm is vulnerable to factors like moving objects, suddenly appeared close obstacles, low illumination, changing lighting, repetitive features, etc. We further verify this in Section \ref{subsec:challenge_case}. 

\subsection{Controlled experiment on challenging cases}
\label{subsec:challenge_case}
One of the key questions is that whether and how much do those challenging features (e.g., day/night, weather, dynamic objects) bother the SLAM algorithms. To demonstrate the effect of those challenging features, we design a controlled experiment, where we collect same trajectories twice, only switching the challenging feature on and off. 

Concretely, we utilize 5 environments (Table \ref{Tab:challenge}): 1) \NAutumnForest{}: w/ and wo/ falling leaves. 2) \NFactory{}: w/ and wo/ moving machinery. 3) \NSoulCity{}: w/ and wo/ rain. 4) \NEndOfWorld{}: outdoor debris w/ and wo/ storm. 5) \NAbandonedFactory{}: day and night. We sample 3-5 trajectories in each environment using medium motion pattern, and run ORB-M and ORB-S on each trajectory for 5 times. 

\begin{figure}[ht]
	\centering
	\includegraphics[width=0.48\textwidth]{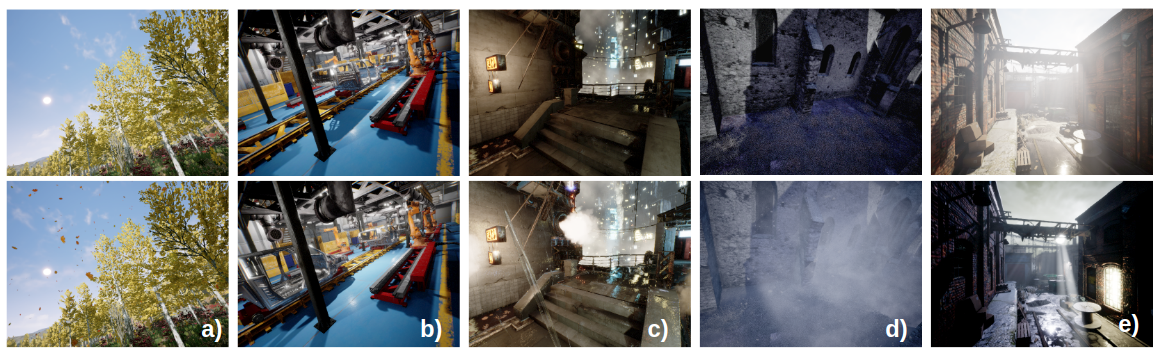}
	\caption{Environments: a) Autumn-forest. b) Factory. c) Soul-city d) End-of-world. e) Abandoned-factory. }
	\label{Fig:challenge_envs}
	\vspace{-0.1in}
\end{figure}

As shown in Table \ref{Tab:challenge}, with challenging features, in 9 out of 10 tests (2 algorithms x 5 environments), the SR or accuracy drops compared to no challenging features. Specifically, we see from the comparison:

1) The dynamic object has limited effect on SR, but remarkably decreases the ATE (3 out of 4 trajectories have a significant drop in ATE). In the \NAutumnForest, the leaves appear in most of the frames but only take small part of the image, while in the \NFactory, the assembly line moves in only a few frames but take large part of the image. The experiment shows the latter has larger impact on the accuracy. 

2) The weather features including rain and storm hurt the SR. ORM-S is more robust to the adverse weather than ORB-M. 
The SR drops more than 50\% with the rain and storm for the ORB-M, and 22\% with storm for the ORB-S. 

3) Low illumination harms SLAM algorithms, but there are limitations in simulating the dark scene. 
We observe that the SLAM algorithm can still extract many features from dark scenes, because of the insufficiency of synthetic data in simulating camera noise or motion blur, which often present in the night scenes in the real world. This would be our future work to investigate image noise models to augment the dark scenes. 

\vspace{-2pt}
\section{Conclusion}
\vspace{-3pt}
We propose TartanAir dataset for visual SLAM in challenging environments. We hope that the proposed dataset and benchmarks will complement others, help reduce overfitting to datasets with limited training or testing examples, and contribute to the development of algorithms that work well in practice. We hope to push the limits of the current visual SLAM algorithms towards real-world applications.  
\vspace{-3pt}
\section*{Acknowledgment}
\vspace{-2pt}
This work was partially supported by Sony award \seqsplit{\#A023367}, ARL award \seqsplit{\#W911NF-P00007}, and ONR award \seqsplit{\#N0014-19-1-2266}. We thank Microsoft for the technical support of AirSim and Azure services. 
         
\begin{table}
	\begin{center}
    \begin{tabular}{l|l|ll|ll}
    \hline
    \multirow{2}{*}{Env} 	 & \multicolumn{1}{c|}{\multirow{2}{*}{Feature}} & \multicolumn{2}{c|}{ORB-M}                         & \multicolumn{2}{c}{ORB-S}                        \\ \cline{3-6} 
                             & \multicolumn{1}{c|}{}                         & \multicolumn{1}{c}{ATE*} & \multicolumn{1}{c|}{SR} & \multicolumn{1}{c}{ATE*} & \multicolumn{1}{c}{SR} \\ \hline
    \multirow{2}{*}{\begin{tabular}[c]{@{}l@{}}Autumn\\ -forest\end{tabular}}     
    		& Static                                        & \textbf{0.093}          & 0.333                   & \textbf{0.087}          & \textbf{0.9}           \\ \cline{2-6} 
            & Dynamic                                       & 0.110                   & 0.333                   & 0.113                   & 0.867                  \\ \hline \hline
    \multirow{2}{*}{Factory}                                                      
    		& Static                                        & 1.108                   & \textbf{1.0}            & \textbf{0.059}          & 1.0                    \\ \cline{2-6} 
            & Dynamic                                       & \textbf{0.786}          & 0.889                   & 0.835                   & 1.0                    \\ \hline \hline
    \multirow{2}{*}{Soul-city}                                                    
    		& No-rain                                       & \textbf{0.165}          & \textbf{0.667}          & 0.401                   & 1.0                    \\ \cline{2-6} 
            & Rain                                          & 0.764                   & 0.375                   & \textbf{0.348}          & 1.0                    \\ \hline \hline
    \multirow{2}{*}{End-of-world}                                                 
    		& No-storm                                      & \textbf{0.106}          & \textbf{0.611}          & \textbf{0.116}          & \textbf{1.0}           \\ \cline{2-6} 
            & Storm                                         & 0.129                   & 0.333                   & 0.292                   & 0.778                  \\ \hline \hline
    \multirow{2}{*}{\begin{tabular}[c]{@{}l@{}}Abandoned-\\ factory\end{tabular}} 
    		& Day                                           & \textbf{0.165}          & \textbf{1.0}            & \textbf{0.0824}         & \textbf{1.0}           \\ \cline{2-6} 
            & Night                                         & 9.385                   & 0.833                   &  0.2407                  & 1.0                    \\ \hline 
    \end{tabular}    \end{center}
	\vspace{-0.1in}
	\caption{Compare the same trajectory w/ and wo/ challenging feature. The best score for each environment is shown in bold font.  *\footnotesize{For a fare comparison of the ATE, we remove failure trajectories on both dynamic and static sides before evaluation}. }
	
	\vspace{-0.2in}
	\label{Tab:challenge}
\end{table}



\bibliographystyle{IEEEtran}
\bibliography{egbib}



\end{document}